\documentclass[graybox]{./styles/svmult}

\usepackage{mathptmx}       
\usepackage{helvet}         
\usepackage{courier}        
\usepackage{type1cm}        
\usepackage{makeidx}         
\usepackage{graphicx}        
\usepackage{multicol}        
\usepackage[bottom]{footmisc}

\usepackage{times}
\usepackage{amsmath}
\usepackage{amssymb}
\usepackage{algorithmic}
\usepackage{algorithm}
\usepackage{subfigure}
\usepackage{multirow}

\makeindex             

\newcommand{\mytextbf}[1]{\vspace{.1cm}\noindent\textbf{{#1}}~}

\usepackage[pagebackref=true,breaklinks=true,letterpaper=true,colorlinks,bookmarks=false]{hyperref}

\begin{document}

\title*{Fast Neighborhood Graph Search using Cartesian Concatenation}

\author{
Jingdong Wang~~
Jing Wang~~
Gang Zeng~~
Rui Gan~~
Shipeng Li~~
Baining Guo}

\institute{Jingdong Wang \at Microsoft Research \email{jingdw@microsoft.com}
\and
Jing Wang,
\at Peking University
\email{cis.wangjing@pku.edu.cn}
\and
Gang Zeng, \at Peking University
\email{g.zeng@ieee.org}
\and
Rui Gan \at Peking University \email{rui_gan@ieee.org}
\and
Shipeng Li
\at Microsoft Research \email{spli@microsoft.com}
\and
Baining Guo
\at Microsoft Research \email{bainguo@microsoft.com}
}

\maketitle

\abstract{
In this paper,
we propose a new data structure
for approximate nearest neighbor search.
This structure
augments the neighborhood graph
with a bridge graph.
We propose to exploit Cartesian concatenation
to produce a large set of vectors,
called bridge vectors,
from several small sets of subvectors.
Each bridge vector is connected with
a few reference vectors near to it,
forming a bridge graph.
Our approach finds nearest neighbors by
simultaneously traversing the neighborhood graph and the bridge graph
in the best-first strategy.
The success of our approach
stems from two factors:
the exact nearest neighbor search over a large number of bridge vectors
can be done quickly,
and the reference vectors connected to a bridge (reference) vector near the query
are also likely to be near the query.
Experimental results
on searching over large scale datasets
(SIFT, GIST and HOG)
show that our approach outperforms state-of-the-art ANN search algorithms
in terms of efficiency and accuracy.
The combination of
our approach
with the IVFADC system~\cite{JegouDS11} also shows
superior performance
over the BIGANN dataset of $1$ billion SIFT features
compared with the best previously published result.
}

\section{Introduction}
Nearest neighbor (NN) search is a fundamental problem
in machine learning, information retrieval
and computational geometry.
It is also a crucial step
in many vision and graphics problems,
such as shape matching~\cite{FromeSSM07},
object retrieval~\cite{PhilbinCISZ07},
feature matching~\cite{BrownL03, SnavelySS06},
texture synthesis~\cite{LiangLXGS01},
image completion~\cite{HaysE07}
and so on.
Recently, the nearest neighbor search problem
attracts more attentions in computer vision
because of the popularity of large scale
and high-dimensional multimedia data.

The simplest solution to NN search is linear scan,
comparing each reference vector to the query vector.
The search complexity is linear with respect to
both the number of reference vectors
and the data dimensionality.
Apparently, it is too time-consuming
and does not scale well in large scale and high-dimensional problems.
Algorithms, including the KD tree~\cite{AryaMNSW98, BeisL97, Bentley75, FriedmanBF77},
BD trees~\cite{AryaMNSW98},
cover tree~\cite{BeygelzimerKL06},
nonlinear embedding~\cite{HwangHA12} and so on,
have been proposed
to improve the search efficiency.
However,
for high-dimensional cases
it turns out
that such approaches are not much more efficient than linear scan
and cannot satisfy the practical requirement.
Therefore,
a lot of efforts have been turned to
approximate nearest neighbor (ANN) search,
such as KD trees with its variants,
hashing algorithms,
neighborhood graph search,
and inverted indices.

In this paper,
we propose a new data structure
for approximate nearest neighbor search~\footnote{A conference version appeared in~\cite{WangWZGLG13}.}.
This structure
augments the neighborhood graph
with a bridge graph
that is able to
boost approximate nearest neighbor search performance.
Inspired by the product quantization technology~\cite{BabenkoL12, JegouDS11},
we adopt Cartesian concatenation (or Cartesian product),
to generate a large set of vectors, which we call \emph{bridge vectors},
from several small sets of subvectors
to approximate the reference vectors.
Each bridge vector is then connected to a few reference vectors
that are near enough to it,
forming a bridge graph.
Combining the bridge graph with the neighborhood graph built over reference data vectors
yields an augmented neighborhood graph.
The ANN search procedure starts by finding the nearest bridge vector to the query vector,
and discovers the first set of reference vectors connected to such a bridge vector.
Then the search simultaneously traverses the bridge graph
and the neighborhood graph
in the best-first manner
using a shared priority queue.

The advantages of adopting the bridge graph
lie in two-fold.
First, computing
the distances
from bridge vectors to the query
is very efficient,
for instance, the computation for $1000000$ bridge vectors
that are formed by $3$ sets of $100$ subvectors
takes almost the same time as that for $100$ vectors.
Second, the best bridge vector is most likely to
be very close to true NNs,
allowing the ANN search to quickly reach true NNs through bridge vectors.

We evaluate the proposed approach
by the feature matching performance
on SIFT and HOG features,
and the performance of searching similar images over tiny images~\cite{TorralbaFF08}
with GIST features.
We show that our approach achieves significant improvements
compared with the state-of-the-art
in terms of accuracy and search time.
We also demonstrate that
our approach in combination with the IVFADC system~\cite{JegouDS11}
outperforms the state-of-the-art
over the BIGANN dataset
of $1$ billion SIFT vectors~\cite{JegouTDA11}.

\section{Literature review}
Nearest neighbor search in the $d$-dimensional metric space $\mathbb{R}^d$
is defined as follows:
given a query $\mathbf{q}$,
the goal is to find an element $\operatorname{NN}(\mathbf{q})$
from the database $\mathcal{X} = \{\mathbf{x}_1, \cdots, \mathbf{x}_n\}$
so that $\operatorname{NN}(\mathbf{q}) = \arg \min_{\mathbf{x} \in \mathcal{X}} \operatorname{dist}(\mathbf{q}, \mathbf{x})$.
In this paper,
we assume that
$\mathbb{R}^d$ is an Euclidean space
and $\operatorname{dist}(\mathbf{q}, \mathbf{x}) = \|\mathbf{q} - \mathbf{x}\|_2$,
which is appropriate for most problems
in multimedia search and computer vision.

There are two types of ANN search problems.
One is error-constrained ANN search
that terminates the search
when the minimum distance found up to now
lies in some scope around the true minimum (or desired) distance.
The other one is time-constrained ANN search
that terminates the search when the search reaches
some prefixed time
(or equivalently examines a fixed number of data points).
The latter category
is shown to be more practical
and give better performance.
Our proposed approach belongs to the latter category.

The ANN search algorithms can be roughly divided
into four categories:
partition trees,
neighborhood graph,
compact codes (hashing and source coding),
and inverted index.
The following presents a short review
of the four categories.

\subsection{Partition trees}
The partition tree based approaches
recursively split the space
into subspaces,
and organize the subspaces via a tree structure.
Most approaches
select hyperplanes or hyperspheres
according to the distribution of data points
to divide the space,
and accordingly data points
are partitioned into subsets.

The KD trees~\cite{Bentley75, FriedmanBF77},
using axis-aligned hyperplane to partition the space,
have been modified
to find ANNs.
Other trees using different partition schemes,
such as BD tress~\cite{AryaMNSW98},
metric trees~\cite{DasguptaF08, LiuMGY04, Moore00, Yianilos93},
hierarchical $k$-means tree~\cite{NisterS06},
and randomized KD trees~\cite{JiaWZZH10, Silpa-AnanH08, WangWJLZZH13},
have been proposed.
FLANN~\cite{MujaL09} aims
to find the best configuration of the hierarchical k-means trees
and randomized KD trees,
and has been shown to work well in practice.

In the query stage,
the branch-and-bound methodology~\cite{Bentley75}
is usually adopted to search (approximate) nearest neighbors.
This scheme needs to traverse the tree
in the depth-first manner
from the root to a leaf
by evaluating the query at each internal node,
and pruning some subtrees
according to the evaluation and the currently-found nearest neighbors.
The current state-of-the-art search strategy,
priority search~\cite{AryaMNSW98} or best-first~\cite{BeisL97},
maintains a priority queue
to access subtrees
in order
so that the data points with large probabilities being true nearest neighbors
are first accessed.
It has been shown that
best-first search (priority search)
achieves the best performance for ANN search,
while the performance might be worse for Exact NN search
than the algorithms without using best-first search.

\subsection{Neighborhood graph search}
The data structure of the neighborhood graph is a directed graph
connecting each vector and its nearest neighbors.
Usually a $R$-NN graph,
that connects each vector to its $R$ nearest neighbors,
is used.
Various algorithms based on neighborhood graph~\cite{AoyamaSSU11, AryaM93b, BeisL97, HajebiASZ11, SameFoun2006, SebastianK02, WangL12}
are developed
for ANN search has been.

The basic procedure of neighborhood graph search
starts from one or several seeding vectors,
and puts them into a priority queue
with the distance to the query being the key.
Then the process proceeds
by popping the top one in the queue,
i.e., the nearest one to the query,
and expanding its neighborhood vectors (from neighborhood graph),
among which the vectors
that have not been visited
are pushed into the priority queue.
This process iterates till
a fixed number of vectors are accessed.

Using neighborhood vectors of a vector
as candidates has two advantages.
One is that extracting the candidates is very cheap
and only takes $O(1)$ time.
The other is that if one vector is close to the query,
its neighborhood vectors are also likely to be close to the query.
The main research efforts
consists of two aspects.
One is to build an effective neighborhood graph~\cite{AoyamaSSU11, SameFoun2006}.
The other is to design efficient and effective ways
to guide the search in the neighborhood graph,
including presetting the seeds created via clustering~\cite{SameFoun2006, SebastianK02},
picking the candidates from KD tress~\cite{AryaM93b},
iteratively searching between KD trees and the neighborhood graph~\cite{WangL12}.
In this paper,
we present a more effective way,
combining the neighborhood graph with a bridge graph,
to search for approximate nearest neighbors.

\subsection{Compact codes}
The compact code approaches
transform each data vector into
a small code,
using the hashing or source coding techniques.
Usually the small code takes much less storage than
the original vector,
and particularly the distance in the small code space,
e.g., hamming distance or using lookup table
can be much more efficiently evaluated
than in the original space.

Locality sensitive hashing (LSH)~\cite{DatarIIM04},
originally used in a manner similar to inverted index,
has been shown to
achieve good theory guarantee
in finding near neighbors with probability,
but it is reported not as good as KD trees
in practice~\cite{MujaL09}.
Multi-probe LSH~\cite{LvJWCL07}
adopts the search algorithm similar to priority search,
achieving a significant improvement.
Nowadays,
the popular usage of hashing is
to use the hamming distance between hash codes
to approximate the distance in the original space
and then adopt linear scan to conduct the search.
To make the best of the data,
recently,
various data-dependent hashing algorithms are proposed
by learning hash functions
using metric learning-like techniques,
including
optimized kernel hashing~\cite{HeLC10},
learned metrics~\cite{JainKG08},
learnt binary reconstruction~\cite{KulisD09},
kernelized LSH~\cite{KulisG09},
and shift kernel hashing~\cite{RaginskyL09},
semi-supervised hashing~\cite{WangKC10A},
(multidimensional) spectral hashing~\cite{WeissFT12, WeissTF08},
spectral hashing~\cite{WeissTF08},
iterative quantization~\cite{GongL11},
complementary hashing~\cite{XuWLZLY11}
and order preserving hashing~\cite{WangWYL13}.

The source coding approach, product quantization~\cite{JegouDS11},
divides the vector into several (e.g., $M$) bands,
and quantizes reference vectors for each band separately.
Then each reference vector is approximated by the nearest center in each band,
and the index for the center is used to represent the reference vector.
Accordingly,
the distance in the original space is approximated
by the distance over the assigned centers in all bands,
which can be quickly computed
using precomputed lookup tables storing the distances between the quantization centers
of each band separately.

\subsection{Inverted index}
Inverted index is composed of
a set of inverted lists
each of which
contains a subset of the reference vectors.
The query stage selects a small number of inverted lists,
regards the vectors contained in the selected inverted lists
as the NN candidates,
and rerank the candidates,
using the distance computed from the original vector
or using the distance computed from the small codes followed
by a second-reranking step
using the distance computed from the original vector,
to find the best candidates.

The inverted index algorithms are widely used for
very large datasets of vectors
(hundreds of million to billions)
due to its small memory cost.
Such algorithms
usually load the inverted index
(and possibly extra codes) into the memory
and store the raw features in the disk.
A typical inverted index is built by clustering algorithms,
e.g.,~\cite{BabenkoL12, JegouDS11, NisterS06, SivicZ09, WangWHL12},
and is composed of a set of inverted lists,
each of which corresponds to a cluster
of reference vectors.
Other inverted indices
include hash tables~\cite{DatarIIM04},
tree codebooks~\cite{Bentley75}
and complementary tree codebooks~\cite{TuPW12}.

\section{Preliminaries}
This section gives short introductions
on several algorithms our approach depends on:
neighborhood graph search,
product quantization,
and the multi-sequence search algorithm.

\subsection{Neighborhood graph search}
A neighborhood graph of a set of vectors $\mathcal{X} = \{\mathbf{x}_1, \cdots, \mathbf{x}_n\}$
is a directed graph
that organizes data vectors
by connecting each data point with its neighboring vectors.
The neighborhood graph is denoted
as $G = \{(v_i, Adj[v_i])\}_{i=1}^n$,
where $v_i$ corresponds to a vector $\mathbf{x}_i$
and $Adj[v_i]$ is a list of nodes that correspond to its neighbors.

The ANN search algorithm proposed in~\cite{AryaM93b},
we call local neighborhood graph search,
is a procedure
that starts from a set of seeding points
as initial NN candidates
and propagates the search
by continuously accessing their neighbors
from previously-discovered NN candidates
to discover more NN candidates.
The~\emph{best-first} strategy~\cite{AryaM93b} is usually adopted
for local neighborhood expansion\footnote{The depth-first search strategy can also be used. Our experiments show that the performance is much worse than the best-first search.}.
To this end,
a~\emph{priority queue} is used to
maintain the previously-discovered NN candidates
whose neighborhoods are not expanded yet,
and initially contains only seeds.
The best candidate
in the priority queue is
extracted out,
and the points in its neighborhood are discovered as new NN candidates
and then pushed into the priority queue.
The resulting search path,
discovering NN candidates,
may not be monotone,
but always attempts to move closer
to the query point without repeating points.
As a local search
that finds better solutions
only from the neighborhood of the current solution,
the local neighborhood graph search will be stuck at a locally optimal point
and has to conduct exhaustive neighborhood expansions
to find better solutions.
Both the proposed approach and the iterated approach~\cite{WangL12} aim
efficiently find solutions beyond local optima.

\subsection{Product quantization}
The idea of product quantization is
to decomposes the space into a Cartesian product
of $M$ low dimensional subspaces
and to quantize each subspace separately.
A vector $\mathbf{x}$ is then
decomposed into $M$ subvectors,
$\mathbf{x}^1, \cdots, \mathbf{x}^M$,
such that $\mathbf{x}^T = [(\mathbf{x}^1)^T~(\mathbf{x}^2)^T \cdots (\mathbf{x}^M)^T]$.
Let the quantization dictionaries
over the $M$ subspaces
be $\mathcal{C}_1, \mathcal{C}_2, \cdots, \mathcal{C}_M$
with $\mathcal{C}_m$ being
a set of centers $\{\mathbf{c}_{m1}, \cdots, \mathbf{c}_{mK}\}$.
A vector $\mathbf{x}$ is represented by a short code
composed of its subspace quantization indices,
$\{k_1, k_2, \cdots, k_M\}$.
Equivalently,
\begin{equation}
\mathbf{x} =
\left[\begin{array}{cccc}
\mathbf{C}^{(1)} & \mathbf{0} & \cdots &\mathbf{0} \\
\mathbf{0} & \mathbf{C}^{(2)} & \cdots &\mathbf{0} \\
\vdots & \vdots & \vdots & \vdots \\
\mathbf{0} & \mathbf{0} & \cdots &\mathbf{C}^{(M)}
\end{array}\right]
\left[\begin{array}{c}
\mathbf{b}^{(1)}\\
\mathbf{b}^{(2)}\\
\vdots\\
\mathbf{b}^{(M)}
\end{array}\right],
\end{equation}
where $\mathbf{b}^{(m)}$ is a vector
in which the $k_m$ entry is $1$
and all others are $0$.

Given a query $\mathbf{q}$,
the asymmetric scheme divides $\mathbf{q}$
into $M$ subvectors $\mathbf{q}^1, \mathbf{q}^M$,
and computes $M$ distance arrays $\{\mathbf{d}_1, \cdots, \mathbf{d}_M\}$
(for computation efficiency, store the square of the Euclidean distance)
with the centers of the $M$ subspaces.
For a database point encoded as $\{k_1, k_2, \cdots, k_M\}$,
the square of the Euclidean distance
is approximated as $\sum_{m=1}^M d_{mk_m}$,
which is called asymmetric distance.

The application of product quantization in our approach
is
different from applications to fast distance computation~\cite{JegouDS11}
and code book construction~\cite{BabenkoL12},
the goal of Cartesian product in this paper
is to build a bridge
to connect the query and the reference vectors
through bridge vectors.

\subsection{Multi-sequence search}
Given several monotonically increasing sequences,
$\{S_b\}_{b=1}^B$
where $S_i$ is a sequence,
$s_b(1), s_b(2), \dots, s_b(L_b)$,
with $s_b(l) < s_b(l+1)$,
the multi-sequence search algorithm~\cite{BabenkoL12}
is able to efficiently traverse the set of $B$-tuples
$\{(s_1(i_1), s_2(i_2), \dots, s_B(i_B)) | i_b = 1 \dots L_b\}$
in order of increasing
the sum $s_1(i_1) + s_2(i_2) + \dots + s_B(i_B)$.

The algorithm uses a min-priority queue
of the tuples $(i_1, i_2, \dots, i_B)$
with the key being the sum $s_1(i_1) + s_2(i_2) + \dots + s_B(i_B)$.
It starts
by initializing the queue with a tuple
$(1, 1, \dots, 1)$.
At step $t$,
the tuple with top priority (the minimum sum),
$(i^{(t)}_1, i^{(t)}_2, \dots, i^{(t)}_B)$,
is popped from the queue
and regarded as the $t$th best tuple whose sum is the $t$th smallest.
At the same time,
the tuple $(i_1, i_2, \dots, i_B)$,
if all its preceding tuples,
$\{(i'_1, i'_2, \dots, i'_B) | i'_b = i_b, i_b-1\} - \{(i_1, i_2, \dots, i_B)\}$
have already been pushed into the queue
is pushed into the queue.
As a result,
the multi-sequence algorithm produces
a sequence of $B$-tuples
in order of increasing the sum
and can stop at step $t-1$
if the best $t$ $B$-tuples are required.
It is shown in~\cite{BabenkoL12} that the time cost of extracting the best $t$ $B$-tuples
is $t\log t$.

\section{Approach}
\label{sec:approach}
The database
$\mathcal{X}$ contains $N$ $d$-dimensional reference vectors,
$\mathcal{X} = \{\mathbf{x}_1, \mathbf{x}_2,
\cdots, \mathbf{x}_N\}$,
$\mathbf{x}_i \in \mathbb{R}^d$.
Our goal is to build an index structure using the bridge graph
such that,
given a query vector $\mathbf{q}$,
its nearest neighbors can be quickly discovered.
In this section,
we first describe the index structure
and then show the search algorithm.

\subsection{Data structure}
\label{sec:approach:datastructure}
Our index structure consists of
two components:
a bridge graph
that connects bridge vectors and their nearest reference vectors,
and a neighborhood graph
that connects
each reference vector to its nearest reference vectors.

\mytextbf{Bridge vectors.}
Cartesian concatenation is an operation
that builds a new set out of a number of given sets.
Given $m$ sets, $\{\mathcal{S}_1, \mathcal{S}_2, \cdots, \mathcal{S}_m\}$,
where each set,
in our case,
contains a set of $d_i$-dimensional subvectors
such that $\sum_{i=1}^m d_i = d$,
the Cartesian concatenation of those sets
is defined as follows,
\begin{align}
\mathcal{Y} =
\times_{i=1}^m \mathcal{S}_i
\triangleq \{\mathbf{y}_j = [\mathbf{y}_{j_1}^T~\mathbf{y}_{j_2}^T~\cdots~\mathbf{y}_{j_m}^T]^T |
\mathbf{y}_{j_i} \in \mathcal{S}_i\}. \nonumber
\end{align}
Here $\mathbf{y}_j$ is a $d$-dimensional vector,
and there exist $\prod_{i=1}^m n_i$ vectors
($n_i = |\mathcal{S}_i|$ is the number of elements in $\mathcal{S}_i$)
in the Cartesian concatenation $\mathcal{Y}$.
Without loss of generality,
we assume that $n_1=n_2=\cdots = n_m = n$ for convenience.
There is a nice property
that identifying
the nearest one
from $\mathcal{Y}$ to a query
only takes $O(dn)$ time
rather than $O(dn^m)$,
despite that the number of elements in $\mathcal{Y}$ is $n^m$.
Inspired by this property,
we use the Cartesian concatenation $\mathcal{Y}$,
called bridge vectors,
as bridges to connect the query vector
with the reference vectors.

\mytextbf{Computing bridge vectors.}
We propose to use product quantization~\cite{JegouDS11},
which aims to minimize the distance of each vector
to the nearest concatenated center derived from subquantizers,
to compute bridge vectors.
This ensures
that the reference vectors discovered through one bridge vector are
not far away from the query
and hence the probability
that those reference vectors
are true NNs is high.

It is also expected
that
the number of reference vectors that are close enough to at least one bridge vector
should be as large as possible
(to make sure
that enough good reference vectors
can be discovered merely
through bridge vectors)
and
that the average number of
the reference vectors discovered through each bridge vector
should be small
(to make sure that the time cost to access them is low).
To this end,
we generate a large amount of bridge vectors.
Such a requirement is similar to~\cite{JegouDS11} for source coding
and different from~\cite{BabenkoL12} for inverted indices.

\mytextbf{Augmented neighborhood graph.}
The augmented neighborhood graph is a combination
of the neighborhood graph $\bar{G}$ over the reference database $\mathcal{X}$
and the bridge graph $B$ between the bridge vectors $\mathcal{Y}$
and the reference vectors $\mathcal{X}$.
The neighborhood graph $\bar{G}$ is a directed graph.
Each node corresponds to a point $\mathbf{x}_i$,
and is also denoted as $\mathbf{x}_i$ for convenience.
Each node $\mathbf{x}_i$ is connected with a list of nodes that correspond to its neighbors,
denoted by$Adj[\mathbf{x}_i]$.

The bridge graph $B$ is constructed
by connecting each bridge vector $\mathbf{y}_j$ in $\mathcal{Y}$
to its nearest vectors $Adj[\mathbf{y}_i]$ in $\mathcal{X}$.
To avoid expensive computation cost,
we build the bridge graph approximately
by finding top $t$ (typically $100$ in our experiments) nearest bridge vectors for each reference vector
and then keeping top $b$ nearest (typically $5$ in our experiments) reference vectors for each bridge vector,
which is efficient and takes $O(Nt(\log t + b))$ time.

The bridge graph is different from the inverted multi-index~\cite{BabenkoL12}.
In the inverted multi-index,
each bridge vector $\mathbf{y}$ contains a list of vectors
that are closer to $\mathbf{y}$ than all other bridge vectors,
while in our approach each bridge is associated with a list of vectors that are closer to $\mathbf{y}$
than all other reference data points.

\subsection{Query the augmented neighborhood graph}
To make the description clear,
without loss of generality,
we assume there are two sets of $n$ subvectors,
$\mathcal{S}_1 = \{\mathbf{y}^1_1, \mathbf{y}^1_2,\cdots, \mathbf{y}^1_n\}$
and
$\mathcal{S}_2 = \{\mathbf{y}^2_1, \mathbf{y}^2_2,\cdots, \mathbf{y}^2_n\}$.
Given a query $\mathbf{q}$ consisting of two subvectors
$\mathbf{q}^1$ and $\mathbf{q}^2$,
the goal is to generate a list of $T$ ($T \ll N$)
candidate reference points from $\mathcal{X}$
where the true NNs of $\mathbf{q}$ are most likely to lie.
This is achieved
by traversing the augmented neighborhood graph
in a best-first strategy.

\begin{figure*}[t]
\centering
\subfigure[Iteration 1]{\includegraphics[scale = .75]{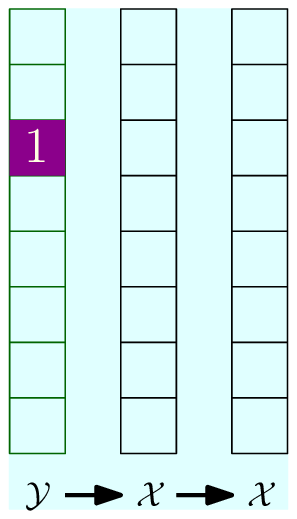}}~~~~~~~~~~
\subfigure[Iteration 2]{\includegraphics[scale = .75]{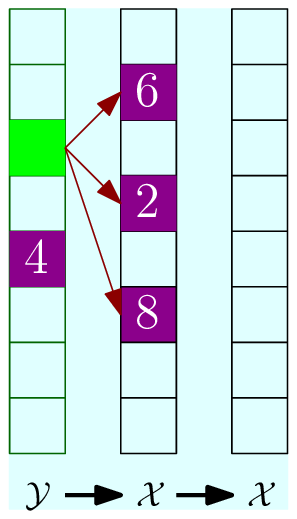}}~~~~~~~~~~
\subfigure[Iteration 3]{\includegraphics[scale = .75]{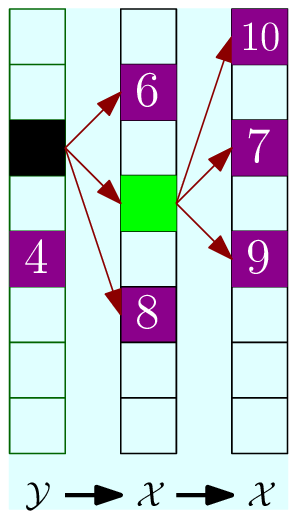}}~~~~~~~~~~
\subfigure[Iteration 4]{\includegraphics[scale = .75]{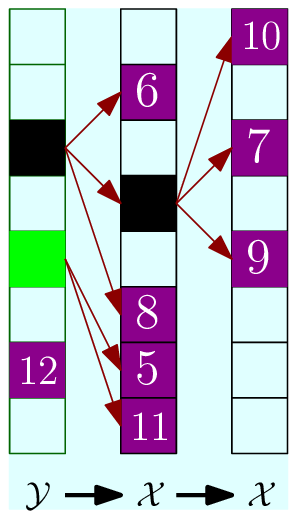}}
\caption{
An example
illustrating the search process.
$\mathcal{Y}\rightarrow \mathcal{X}$: the bridge graph,
and $\mathcal{X}\rightarrow \mathcal{X}$: the neighborhood graph.
The white numbers are the distances to the query.
Magenta denotes the vectors in the main queue,
green represents the vector being popped out from the main queue,
and black indicates the vectors whose neighborhoods have already been expanded
}
\label{fig:ExampleSearchProcess}
\end{figure*}

We give a brief overview of the ANN search procedure over a neighborhood graph
before describing how to make use of bridge vectors.
The algorithm begins with a set of (one or several) vectors $\mathcal{P}_s = \{\mathbf{p}\}$
that are contained in the neighborhood graph.
It maintains a set of nearest neighbor candidates
(whose neighborhoods have not been expanded),
using a min-priority queue,
which we call the main queue,
with the distance to the query as the key.
The main queue initially contains the vectors in $\mathcal{P}_s$.
The algorithm proceeds by iteratively expanding the neighborhoods
in a best-first strategy.
At each step,
the vector $\mathbf{p}^*$ with top priority
(the nearest one to $\mathbf{q}$) is popped from the queue.
Then each neighborhood vector in $Adj[\mathbf{p}^*]$
is inserted to the queue if it is not visited,
and at the same time
it is added to the result set
(maintained by a max-priority queue
with a fixed length depending on how many nearest neighbors
are expected).

To exploit the bridge vectors,
we present an extraction-on-demand strategy,
instead of fetching all the bridge vectors
to the main queue,
which leads to expensive cost
in sorting them and maintaining the main queue.
Our strategy is to maintain the main queue
such that it consists of only one bridge vector
if available.
To be specific,
if the top vector $\mathbf{p}^*$ in the main queue
is a reference vector,
the algorithm proceeds
as usual,
the same to the above procedure without using bridge vectors.
If the top vector is a bridge vector,
we first insert its neighbors $Adj[\mathbf{p}^*]$
into the main queue and the result set,
and in addition we find the next nearest bridge vector (to the query $\mathbf{q}$)
and insert it to the main queue.
The pseudo code of the search algorithm
is given in Algorithm~\ref{algorithm:CoordinateWiseEnumeration}
and an example process is illustrated
in Figure~\ref{fig:ExampleSearchProcess}.

Before traversing the augmented neighborhood graph,
we first process the bridge vectors,
and compute the distances
(the square of the Euclidean distance)
from $\mathbf{q}^1$ to the subvectors in $\mathcal{S}_1$
and from $\mathbf{q}^2$ to the subvectors in $\mathcal{S}_2$,
and then sort the subvectors in the order of increasing distances,
respectively.
We denote the sorted subvectors
as $\{\mathbf{y}^1_{i_1},\cdots, \mathbf{y}^1_{i_n}\}$
and $\{\mathbf{y}^2_{j_1},\cdots, \mathbf{y}^2_{j_n}\}$.
As the size $n$ of $\mathcal{S}_1$ and $\mathcal{S}_2$
is typically not large
(e.g., $100$ in our case),
the computation cost is very small
(See details in Section~\ref{sec:discussion}).

The extraction-on-demand strategy
needs to visit the bridge vector
one by one
in the order of increasing distance from $\mathbf{q}$.
It is easily shown that
$\operatorname{dist}^2(\mathbf{q}, \mathbf{y})
= \operatorname{dist}^2(\mathbf{q}^1, \mathbf{y}^1) +
\operatorname{dist}^2(\mathbf{q}^2, \mathbf{y}^2)$,
where $\mathbf{y}$ is consists of $\mathbf{y}^1$ and $\mathbf{y}^2$.
Naturally, $\mathbf{y}_{i_1, j_1}$,
composed of $\mathbf{y}^1_{i_1}$ and $\mathbf{y}^2_{i_1}$,
is the nearest one to $\mathbf{q}$.
The multi-sequence algorithm
(corresponding to ExtractNextNearestBridgeVector()
in Algorithm~\ref{algorithm:CoordinateWiseEnumeration})
is able to
fast produce a sequence of pairs $(i_k, j_l)$
so that the corresponding bridge vectors
are visited in the order of increasing distances to the query $\mathbf{q}$.
The algorithm is very efficient
and producing the $t$-th bridge vector only takes $O(\log(t))$ time.
Slightly different from extracting a fixed number of
nearest bridge vectors once~\cite{BabenkoL12},
our algorithm automatically determines when to extract the next one,
that is when there is no bridge vector in the main queue.

\floatname{algorithm}{\small{Algorithm}}
\begin{algorithm}[t]
\algsetup{
   linenosize=\small,
   linenodelimiter=.
}
\caption{\small{ANN search over the augmented neighborhood graph}}
\label{algorithm:CoordinateWiseEnumeration}
\begin{algorithmic}[1]
\small
\item[~] {/* $\mathbf{q}$: the query;
$\mathcal{X}$: the reference data vectors;
$\mathcal{Y}$: the set of bridge vectors;
$G$: the augmented neighborhood graph;
$Q$: the main queue;
$R$: the result set;
$T$: the maximum number of discovered vectors; */}
\item[\textbf{Procedure}]
ANNSearch($\mathbf{q}$, $\mathcal{X}$, $\mathcal{Y}$, $G$, $Q$, $R$, $T$)
\STATE \emph{/* Mark each reference vector undiscovered */}
\FOR {each $\mathbf{x} \in \mathcal{X}$}
    \STATE Color[$\mathbf{x}$] $\gets$ white;
\ENDFOR
\STATE \emph{/* Extract the nearest bridge vector */}
\STATE $(\mathbf{y}, D) \gets$ ExtractNextNearestBridgeVector($\mathcal{Y}$);
\STATE $Q \gets (\mathbf{y}, D)$;
\STATE $t \gets 0$

\STATE \emph{/* Start the search */}
\WHILE {($Q \neq \emptyset$ \&\& $t \leqslant T$)}
    \STATE \emph{/* Pop out the best candidate vector and expand its neighbors */}
    \STATE $(\mathbf{p}, D) \gets Q.\operatorname{pop}$();
    \FOR {each $\mathbf{x} \in Adj[\mathbf{p}]$}
        \IF {Color[$\mathbf{x}$] = white}
            \STATE $D \gets \operatorname{dist}(\mathbf{q}, \mathbf{x})$;
            \STATE $Q \gets (\mathbf{x}, D)$;
            \STATE Color[$\mathbf{x}$] $\gets$ black;~\emph{/* Mark it discovered */}
            \STATE $R \gets (\mathbf{x}$, $D$);~\emph{/* Update the result set */}
            \STATE $t \gets t + 1$;
        \ENDIF
    \ENDFOR
    \STATE \emph{/* Extract the next nearest bridge vector if $\mathbf{p}$ is a bridge vector */}
    \IF {$\mathbf{p} \in \mathcal{Y}$}
        \STATE $(\mathbf{y}, D) \gets$ ExtractNextNearestBridgeVector($\mathcal{Y}$);
        \STATE $Q \gets (\mathbf{y}, D)$;
    \ENDIF
\ENDWHILE
\RETURN $R$;
\end{algorithmic}
\end{algorithm}

\section{Experiments}
\subsection{Setup}
We perform our experiments
on three large datasets:
the first one with local SIFT features,
the second one with global GIST features,
and the third one with HOG features,
and a very large dataset,
the BIGANN dataset of $1$ billion SIFT features~\cite{JegouTDA11}.

The SIFT features are collected
from the Caltech $101$ dataset~\cite{FeiFP04}.
We extract maximally stable extremal regions (MSERs)
for each image,
and compute a $128$-dimensional byte-valued SIFT feature for each MSER.
We randomly sample $1000K$ SIFT features
and $100K$ SIFT features,
respectively as the reference and query set.
The GIST features are extracted
on the tiny image set~\cite{TorralbaFF08}.
The GIST descriptor is a $384$-dimensional byte-valued vector.
We sample $1000K$ images as the reference set
and $100K$ images as the queries.
The HOG descriptors are extracted
from Flickr images,
and each HOG descriptor is
a $512$-dimensional byte-valued vector.
We sample $10M$ HOG descriptors as the reference set
and $100K$ as the queries.
The BIGANN dataset~\cite{JegouTDA11} consists of $1B$ $128$-dimensional byte-valued vectors
as the reference set
and $10K$ vectors as the queries.

We use the accuracy score to evaluate the search quality.
For $k$-ANN search,
the accuracy is computed
as $r/k$, where $r$ is the number of retrieved vectors
that are contained in the true $k$ nearest neighbors.
The true nearest neighbors are computed
by comparing each query with all the reference vectors
in the data set.
We compare different algorithms
by calculating the search accuracy
given the same search time,
where the search time is recorded
by varying the number of accessed vectors.
We report the performance
in terms of search time vs. search accuracy
for the first three datasets.
Those results are obtained
with $64$ bit programs on
a $3.4G$ Hz quad core Intel PC
with $24G$ memory.

\begin{table}[t]
\centering
\caption{The parameters of our approach
and the statistics.
\#reference means the number of reference vectors
associated with the bridge vectors,
and $\alpha$
means the average number of unique reference vectors
associated with each bridge vector}
\label{tab:parameter}
\begin{tabular}[b]{|@{~~}c@{~~}||@{~~}c@{~~}|@{~~}c@{~~}|@{~~}c@{~~}||@{~~}c@{~~}|@{~~}c@{~~}|}
\hline
 & size & \#partitions & \#clusters & \#reference& $\alpha$ \\
\hline \hline
SIFT & $1M$ & $4$ &  $50$ & $715K$ & $11.4\%$\\
GIST & $1M$ & $4$ &  $50$ & $599K$ & $9.59\%$ \\
HOG & $10M$ & $4$ &  $100$ & $5730K$ & $5.73\%$\\
\hline
\end{tabular}
\end{table}

\subsection{Empirical analysis}
\begin{figure}[t]
\centering
\subfigure[]
{\includegraphics[height = .35\linewidth,clip]{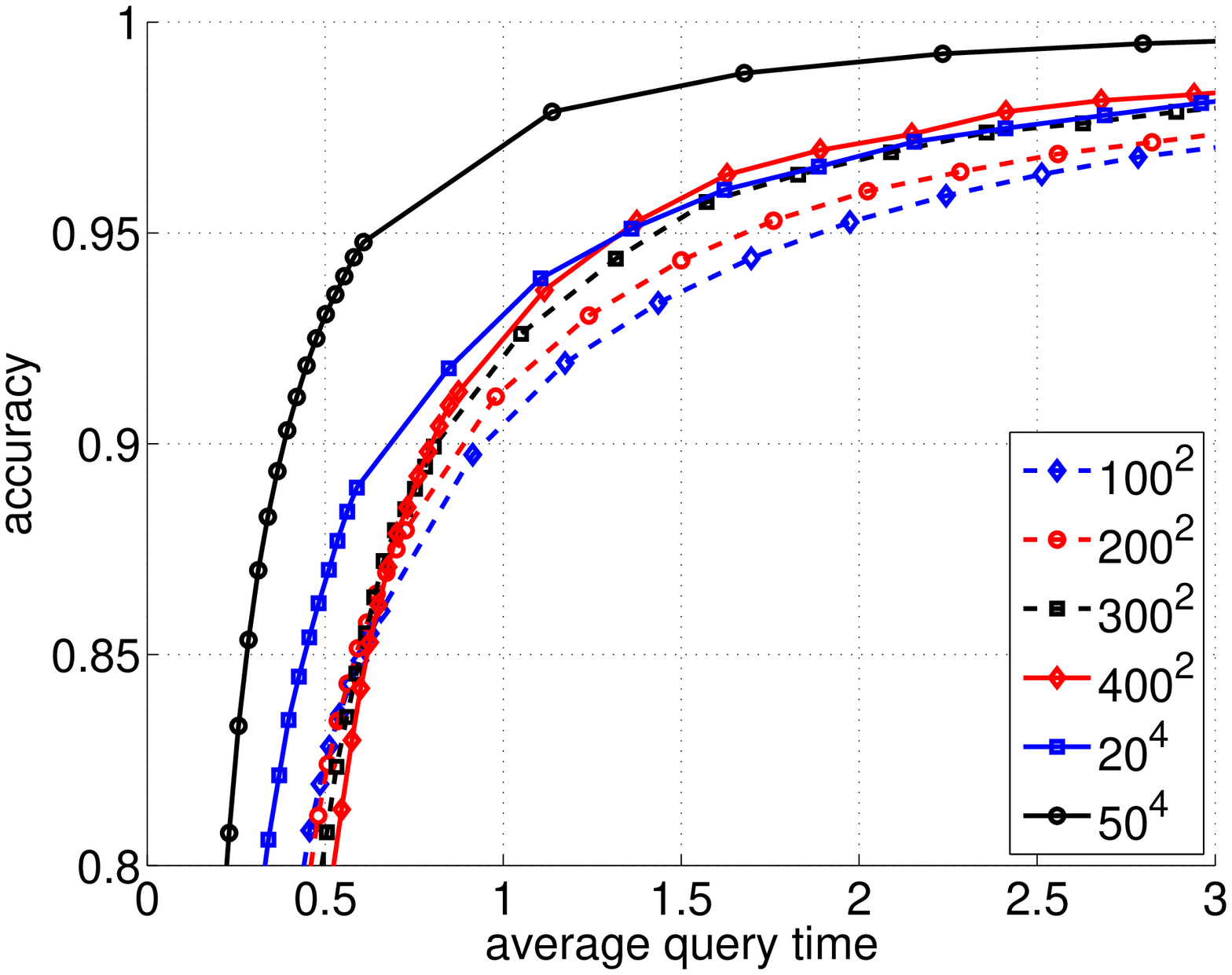}}~~~~~~~
\subfigure[]
{\includegraphics[height = .35\linewidth,clip]{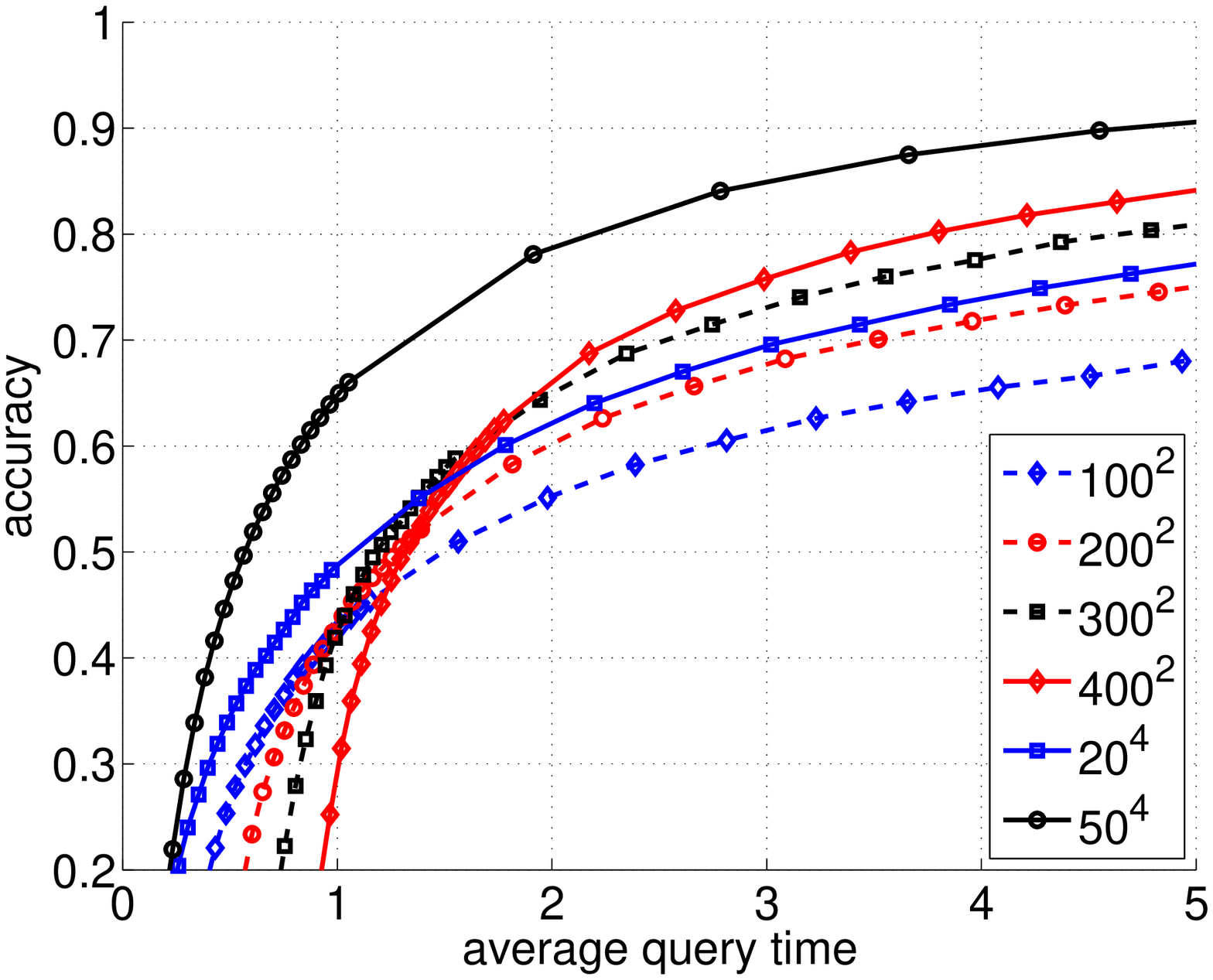}}
\caption{Search performances with different number of partitions
and clusters over (a) $1M$ SIFT and (b) $1M$ GIST.
$x^y$: $y$ means \#partitions and $x$ is \#clusters
 }\label{fig:PartitionsAndClusters}
\end{figure}

The index structure construction in our approach
includes partitioning the vector into $m$ subvectors
and grouping the vectors of each partition into $n$ clusters.
We conduct experiments to
study how they influence the search performance.
The results over the $1M$ SIFT and $1M$ GIST datasets
are shown in Figure~\ref{fig:PartitionsAndClusters}.
Considering two partitions,
it can be observed that
the performance becomes better
with more clusters for each partition.
This is
because more clusters produce more bridge vectors
and thus more reference vectors are associated with bridge vectors
and their distances are much smaller.
The result with $4$ partitions and $50$ clusters per partition
gets the best performance
as in this case the properties desired for bridge vectors
described in Section~\ref{sec:approach:datastructure}
are more likely to be satisfied.

\subsection{Comparisons}
We compare our approach
with state-of-the-art algorithms,
including
iterative neighborhood graph search~\cite{WangL12},
original neighborhood graph search (AryaM93)~\cite{AryaM93b},
trinary projection (TP) trees~\cite{JiaWZZH10},
vantage point (VP) tree~\cite{Yianilos93},
Spill trees~\cite{LiuMGY04},
FLANN~\cite{MujaL09},
and inverted multi-index~\cite{BabenkoL12}.
The results of all other methods are obtained
by well tuning parameters.
We do not report the results from hashing algorithms
as they are much worse than tree-based approach,
which is also reported in~\cite{MujaL09, WangWJLZZH13}.
The neighborhood graphs of different algorithms
are the same,
and each vector is connected with $20$ nearest vectors.
We construct approximate neighborhood graphs
using the algorithm~\cite{WangWZTGL12}.
Table~\ref{tab:parameter}
shows the parameters for our approach,
together with some statistics.

\begin{figure*}[ht]
\renewcommand{\thesubfigure}{}
\makeatletter
\renewcommand{\@thesubfigure}{}
\centering
~~~~\includegraphics[width = .85\linewidth, clip]{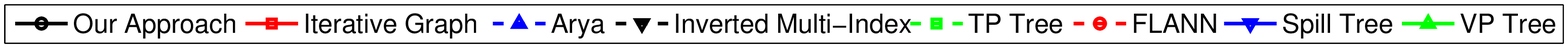}\\
\vspace{.1cm}
\scriptsize{(a)}
\hspace{0.1cm}
\subfigure
{\label{fig:result:101_1nn}
\includegraphics[width=0.26\linewidth, clip]{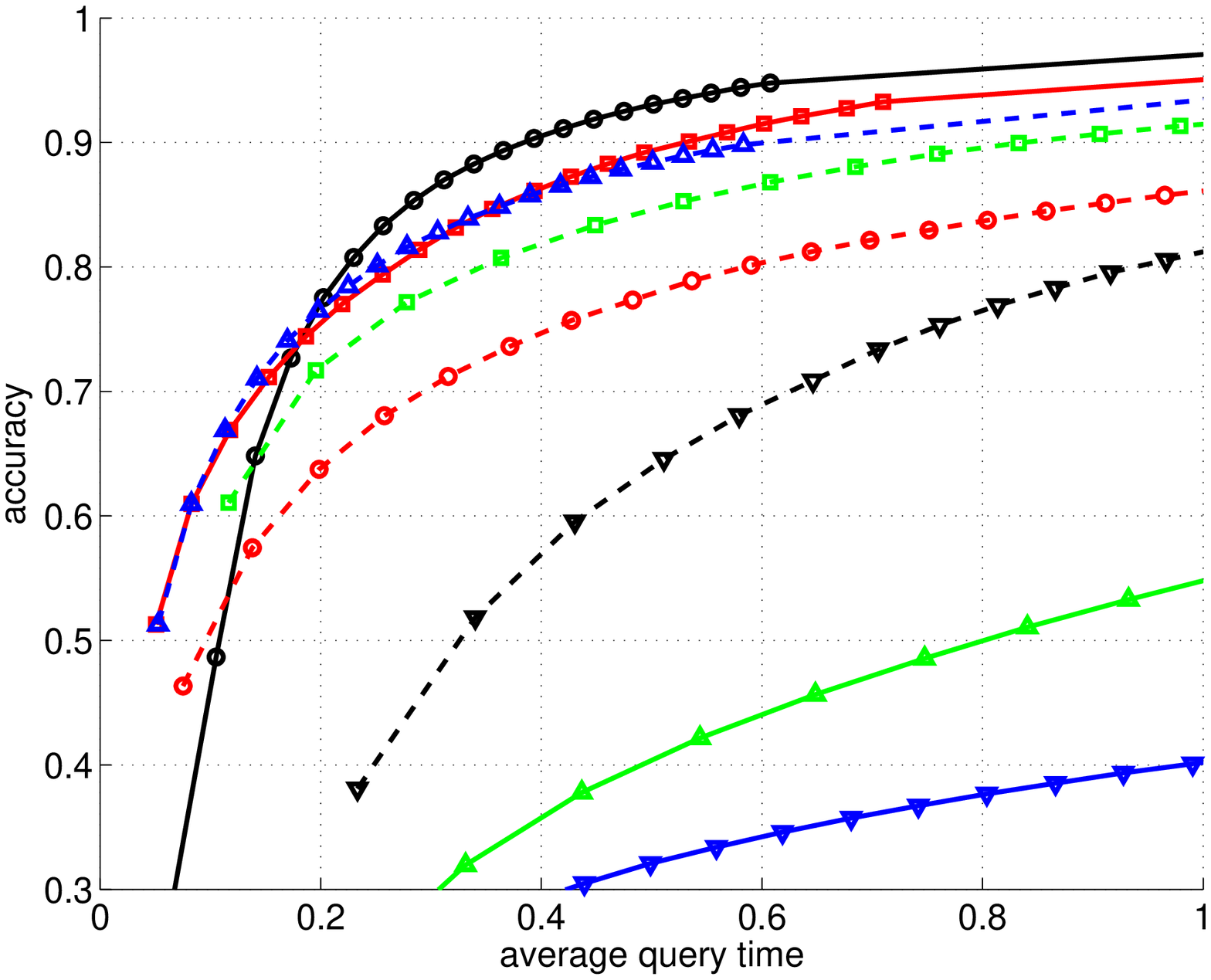}}~~~~~~~~~
\subfigure
{\label{fig:result:101_10nn}
\includegraphics[width=0.26\linewidth, clip]{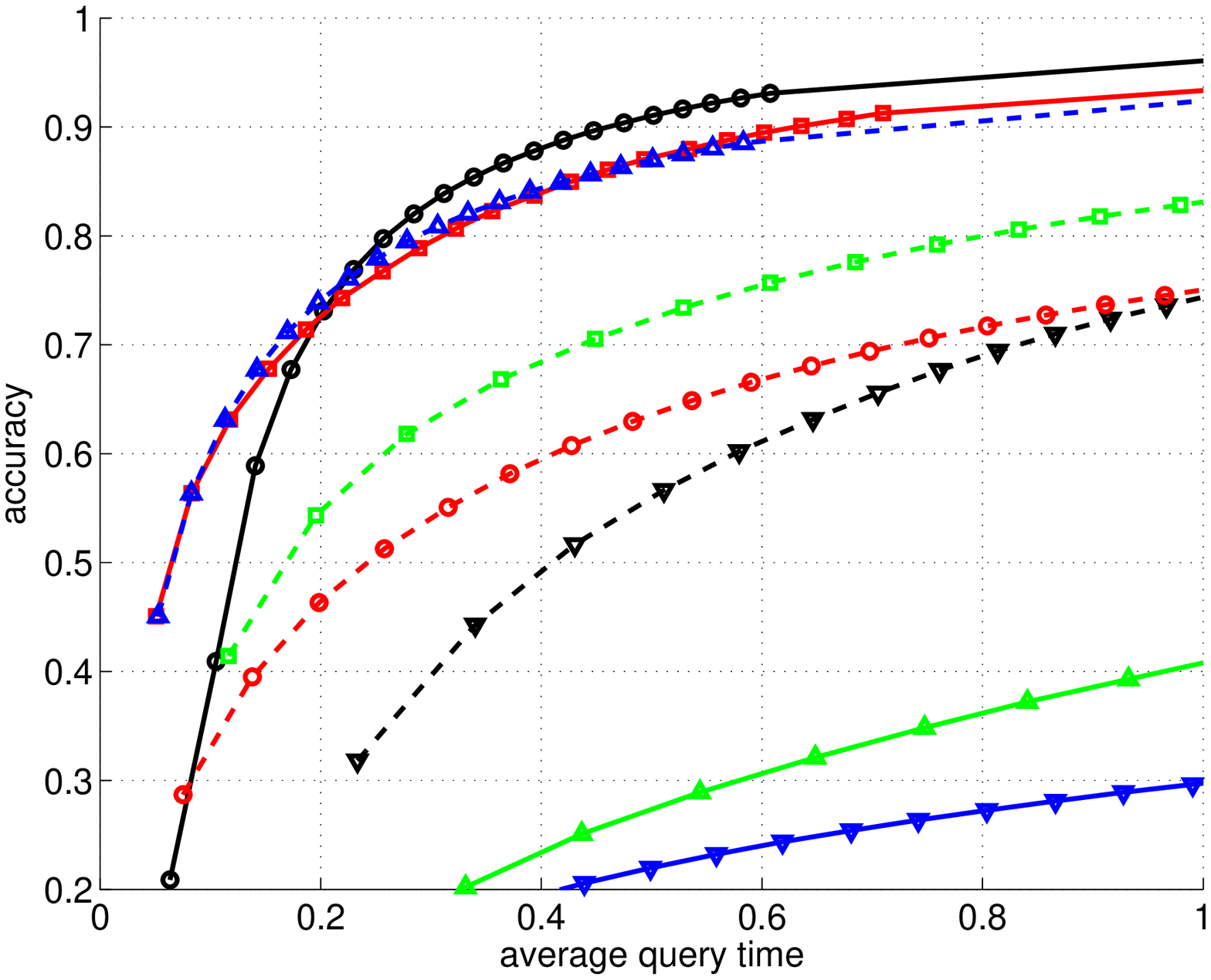}}~~~~~~~~~
\subfigure
{\label{fig:result:101_100nn}
\includegraphics[width=0.26\linewidth, clip]{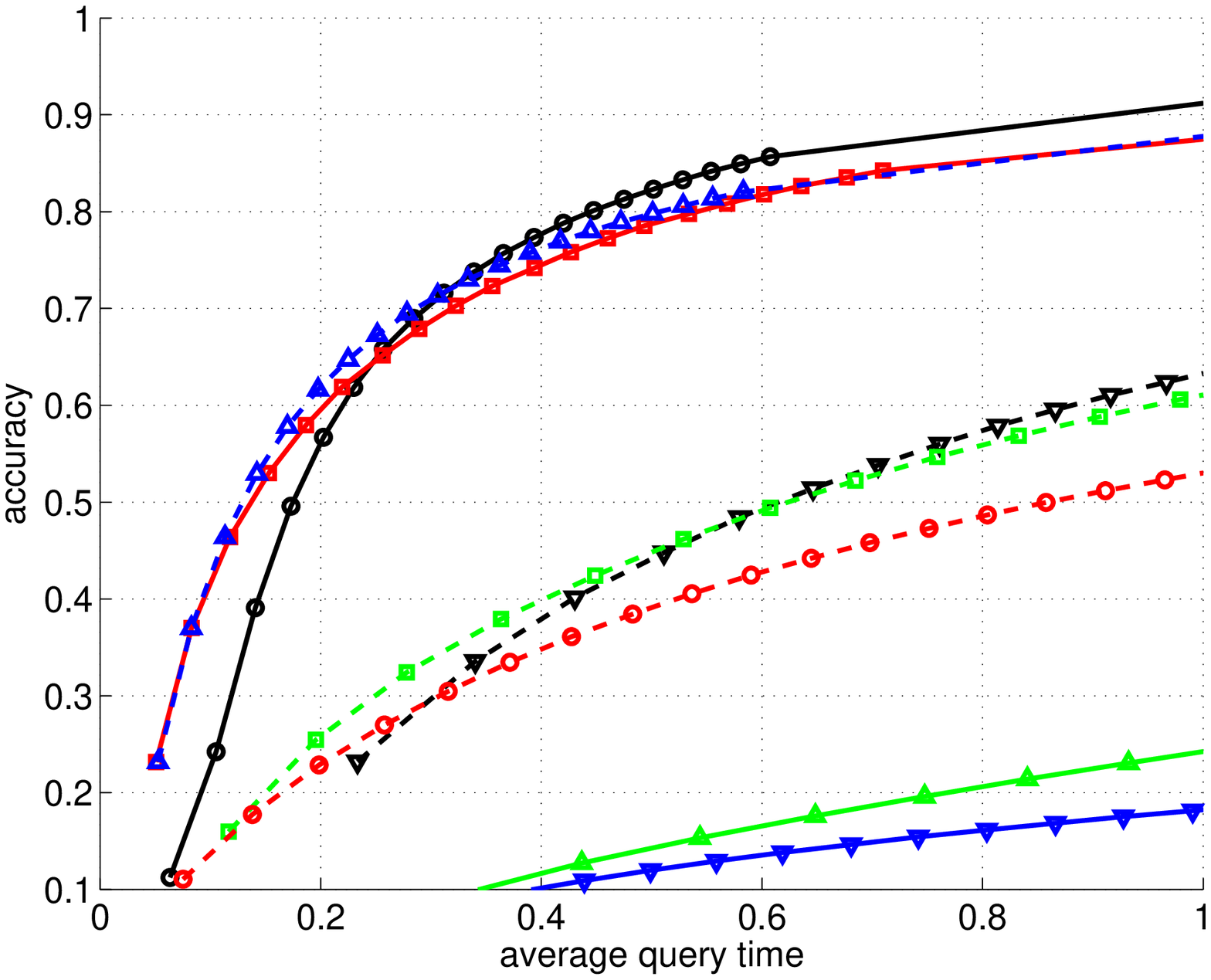}}\\
\scriptsize{(b)}
\hspace{0.1cm}
\subfigure
{\label{fig:result:tiny_1nn}
\includegraphics[width=0.26\linewidth, clip]{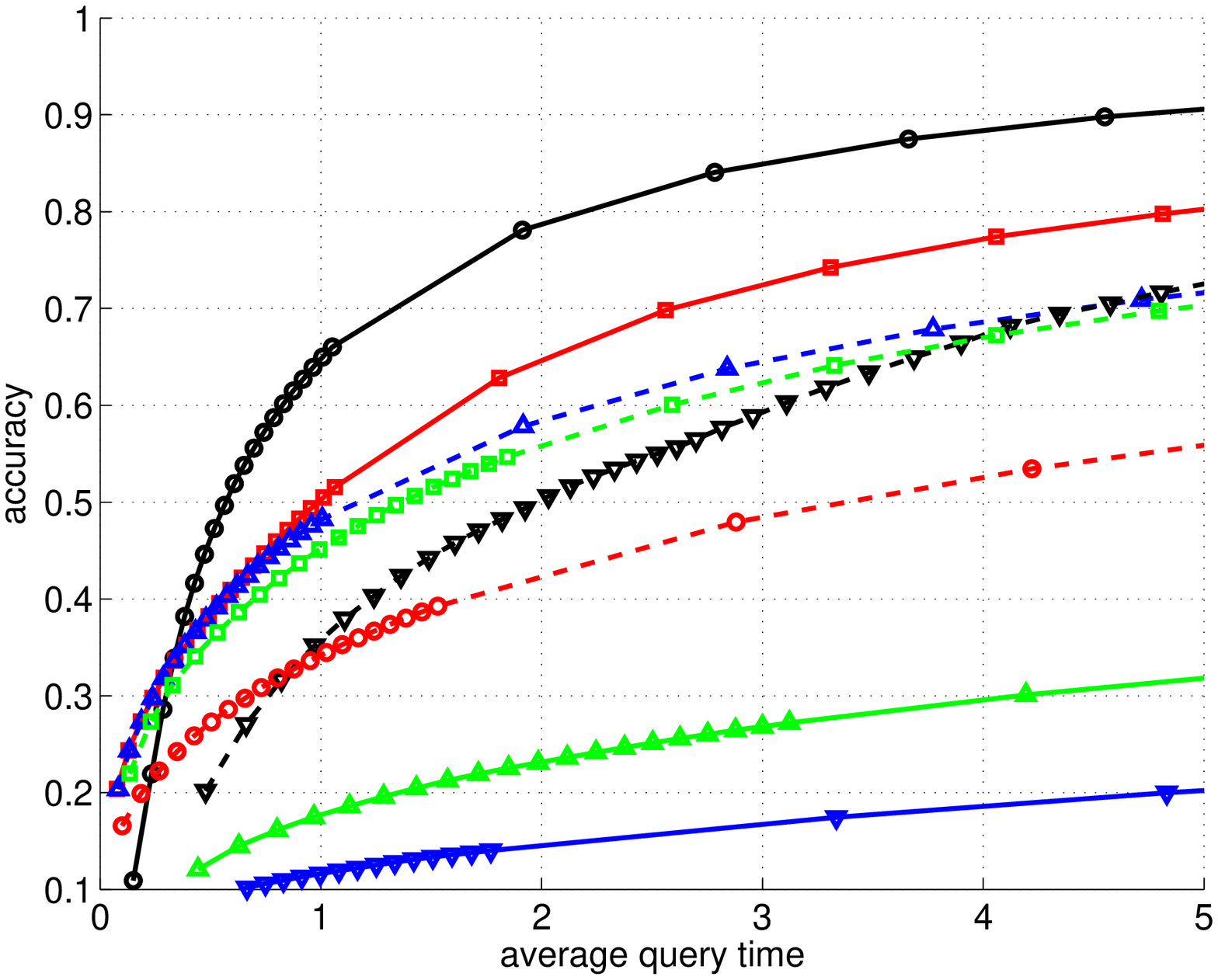}}~~~~~~~~~
\subfigure
{\label{fig:result:tiny_10nn}
\includegraphics[width=0.26\linewidth, clip]{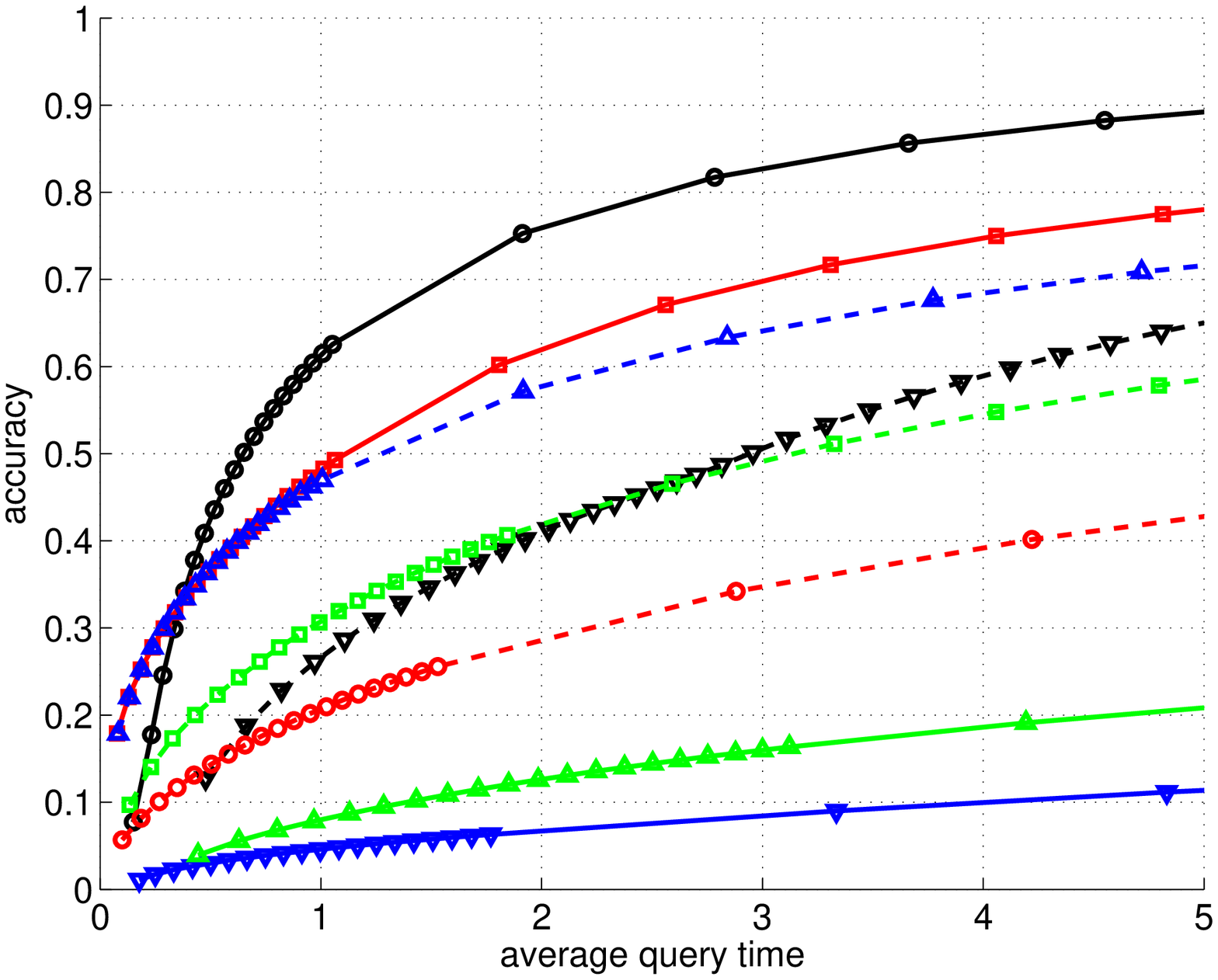}}~~~~~~~~~
\subfigure
{\label{fig:result:tiny_100nn}
\includegraphics[width=0.26\linewidth, clip]{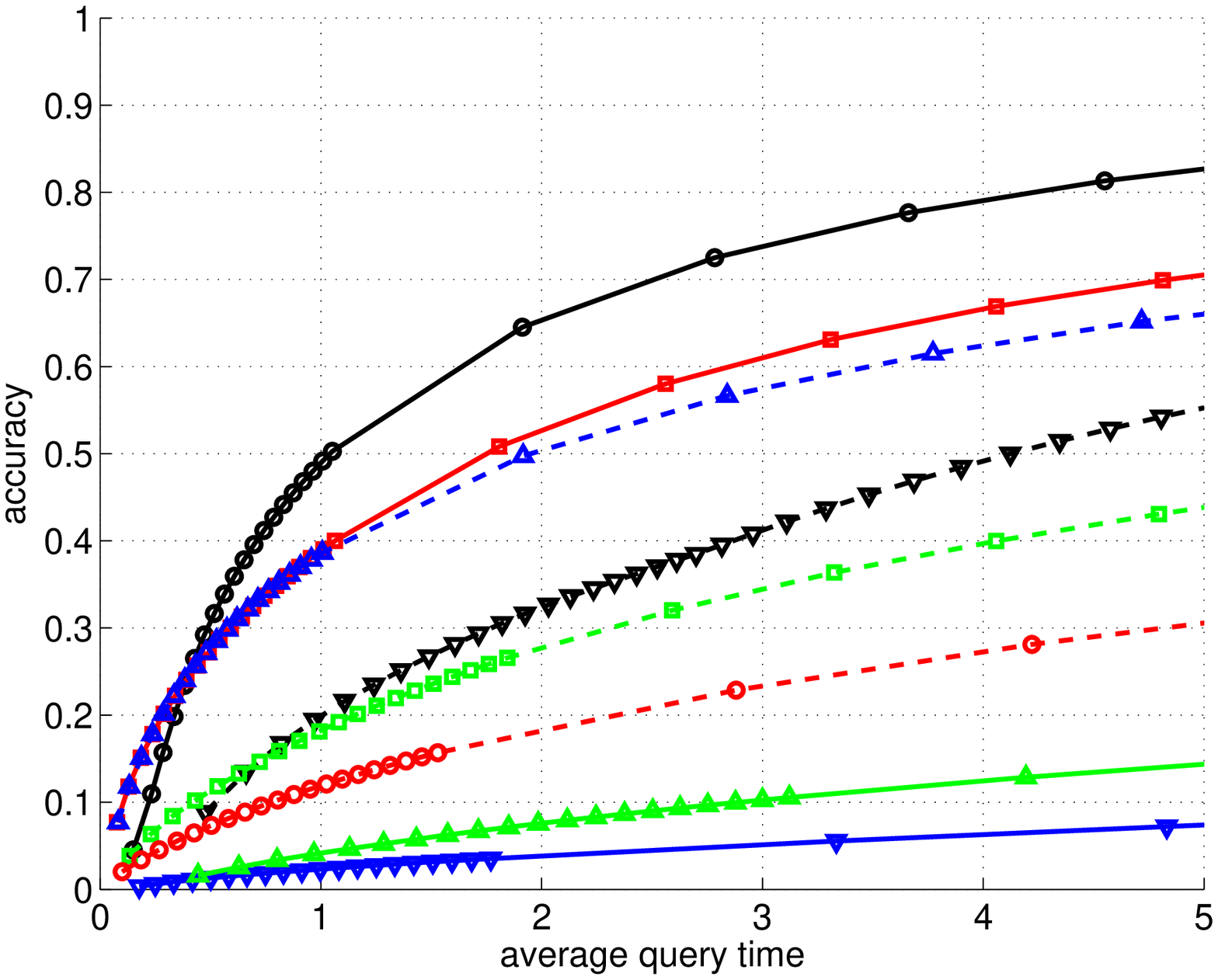}}\\
\scriptsize{(c)}
\subfigure[$k = 1$]
{\label{fig:result:tsv_1nn}
\includegraphics[width=0.26\linewidth, clip]{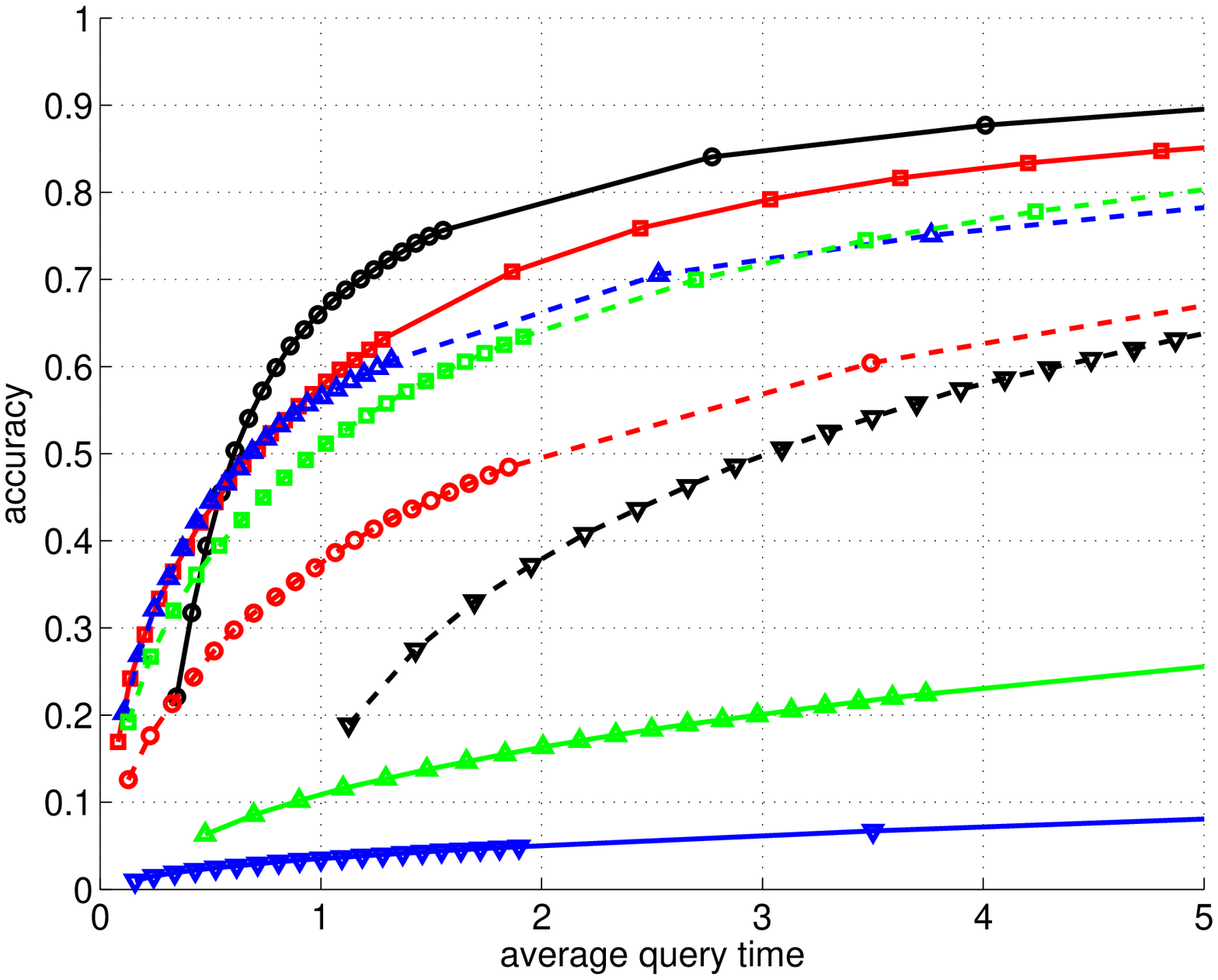}}~~~~~~~~~
\subfigure[$k = 10$]
{\label{fig:result:tsv_10nn}
\includegraphics[width=0.26\linewidth, clip]{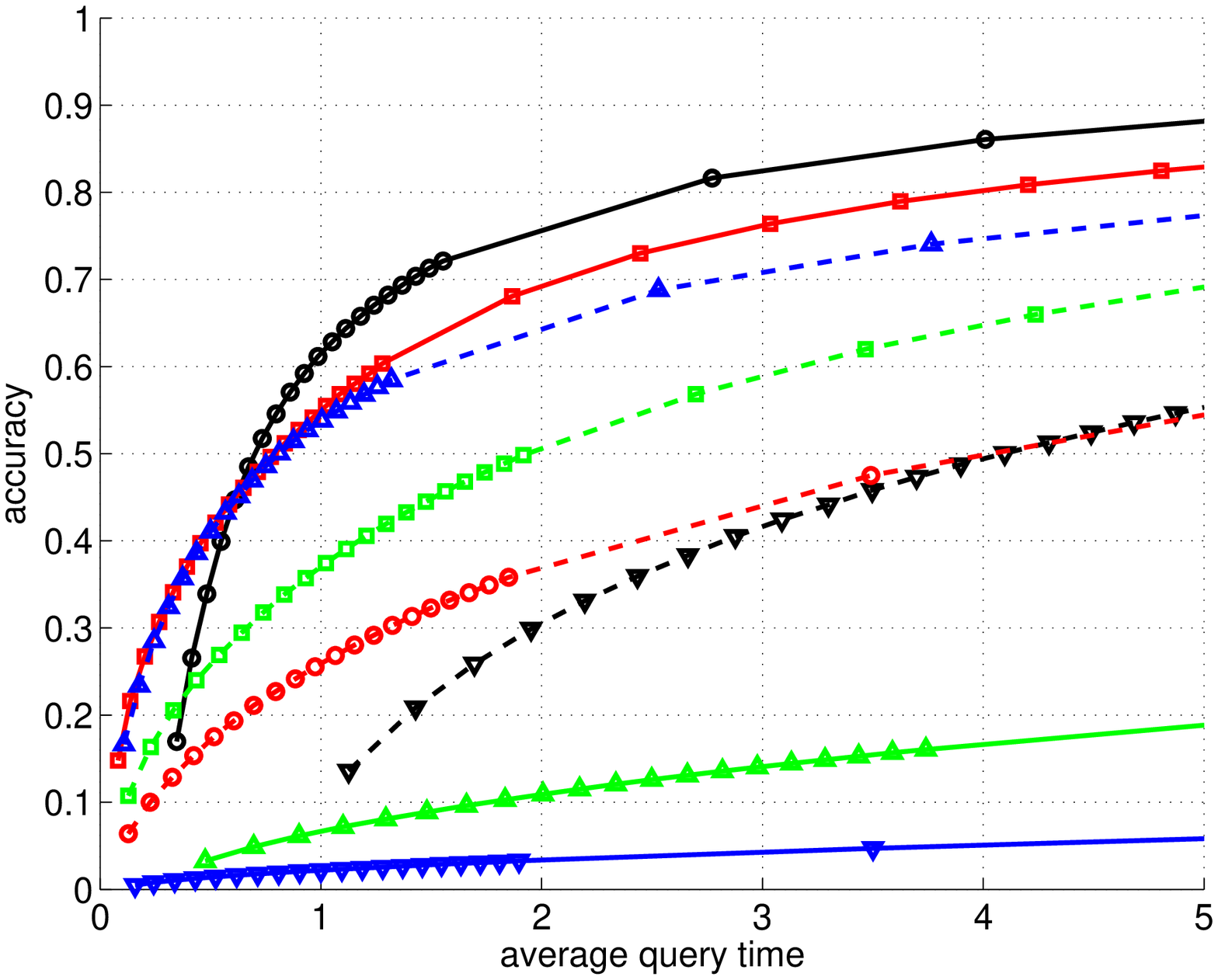}}~~~~~~~~~
\subfigure[$k = 100$]
{\label{fig:result:tsv_100nn}
\includegraphics[width=0.26\linewidth, clip]{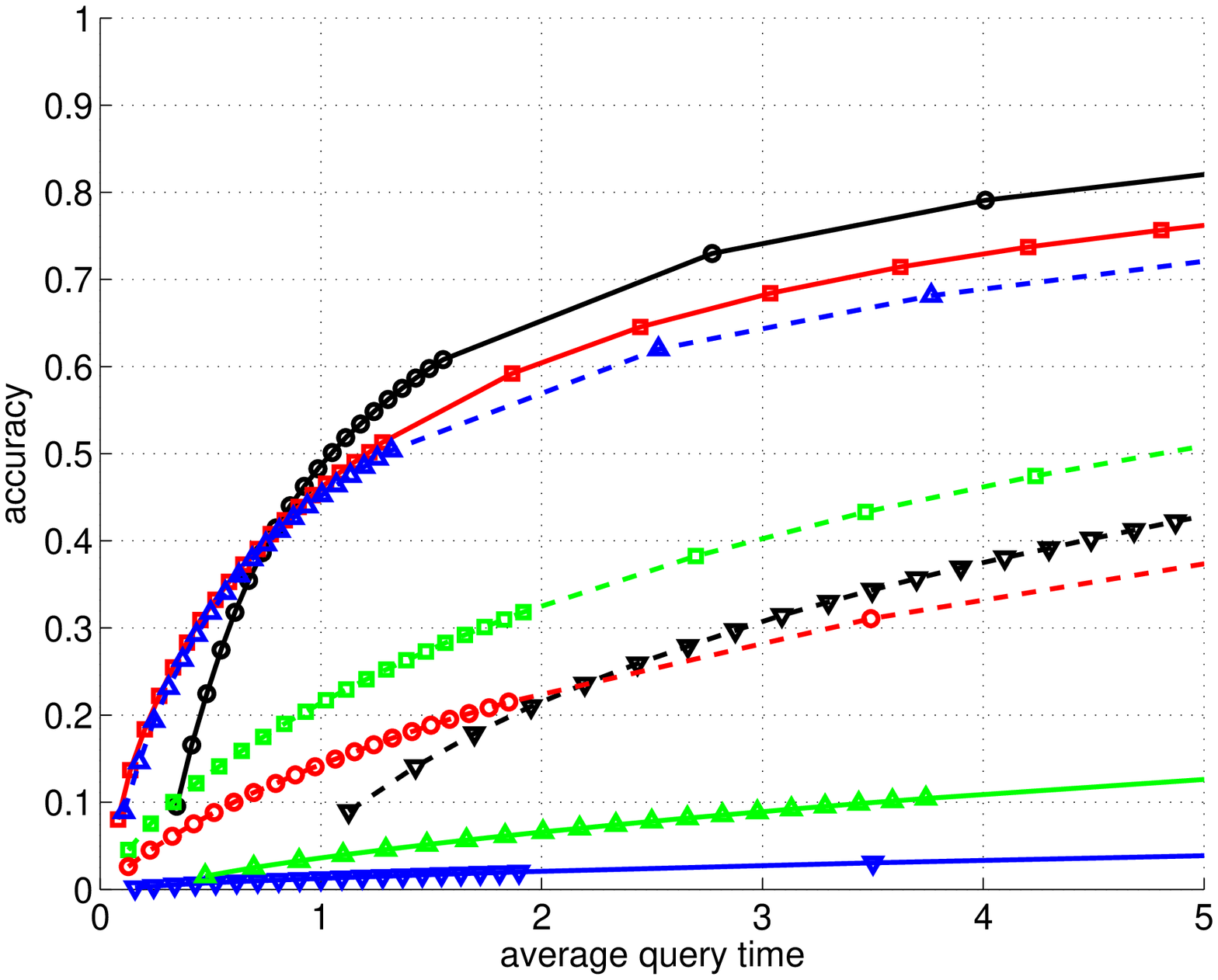}}\\
\centering
\caption{Performance comparison on
(a) $1M$ $128$-dimensional SIFT features,
(b) $1M$ $384$-dimensional GIST features,
and (c) $10M$ $512$-dimensional HOG features.
$k$ is the number of target nearest neighbors}
\label{fig:result}
\end{figure*}

The experimental comparisons are shown in Figure~\ref{fig:result}.
The horizontal axis corresponds to search time (milliseconds),
and the vertical axis corresponds to search accuracy.
From the results over the SIFT dataset
shown in the first row of Figure~\ref{fig:result},
our approach performs the best.
We can see that,
given the target accuracy $90\%$
$1$-NN and $10$-NN,
our approach
takes about $\frac{2}{3}$ time
of the second best algorithm,
iterative neighborhood graph search.

The second row of Figure~\ref{fig:result} shows the results over the GIST dataset.
Compared with the SIFT feature (a $128$-dimensional vector),
the dimension of the GIST feature ($384$) is larger
and the search is hence more challenging.
It can be observed that
our approach is still consistently
better than other approaches.
In particular,
the improvement is more significant,
and for the target precision $70\%$
our approach takes only half time
of the second best approach,
from $1$ to $100$ NNs.
The third row of Figure~\ref{fig:result}
shows the results over the HOG dataset.
This data set is the most difficult
because it contains more ($10M$) descriptors
and its dimension is the largest ($512$).
Again, our approach achieves the best results.
For the target accuracy $70\%$,
the search time in the case of $1$ NN is about
$\frac{4}{7}$ of the time of the second best algorithm.

All the neighborhood graph search algorithms
outperform the other algorithms,
which shows that
the neighborhood graph structure is good
to index vectors.
The superiority of our approach to
previous neighborhood graph algorithms
stems from
that our approach exploits the bridge graph
to help the search.
Inverted multi-index does not produce competitive results
because its advantage is small index structure size but its search performance is
limited by an unfavorable trade-off between the search accuracy
and the time overhead in quantization.
It is shown in~\cite{BabenkoL12}
that inverted multi-index
works the best when using a second-order multi-index
and a large codebook,
but this results in high quantization cost.
In contrast,
our approach benefits from
the neighborhood graph structure
so that we can use a high-order product quantizer
to save the quantization cost.

\begin{figure}[t]
\centering
\subfigure[]
{\includegraphics[width = .45\linewidth,clip]{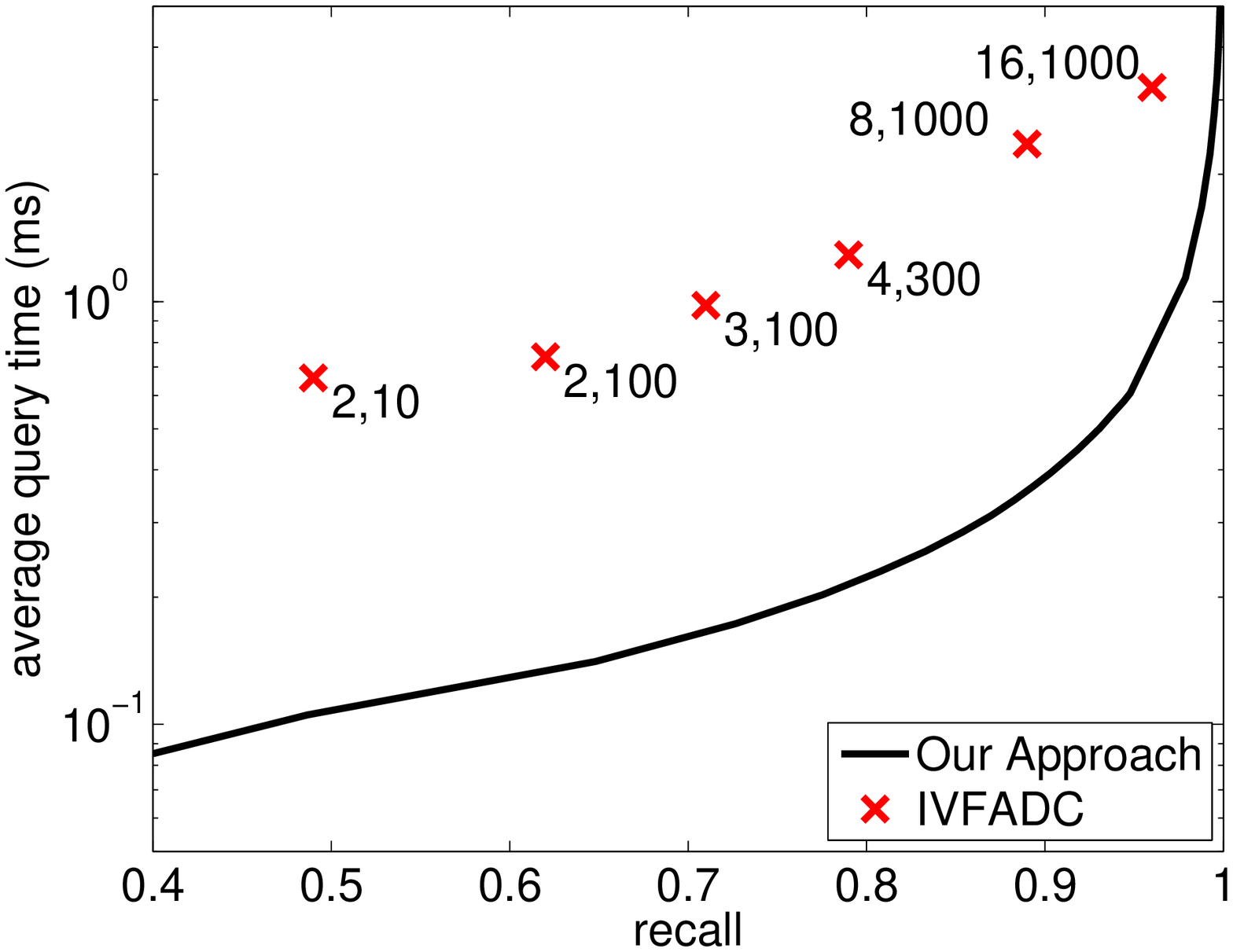}}~~~~~~
\subfigure[]
{\includegraphics[width = .45\linewidth,clip]{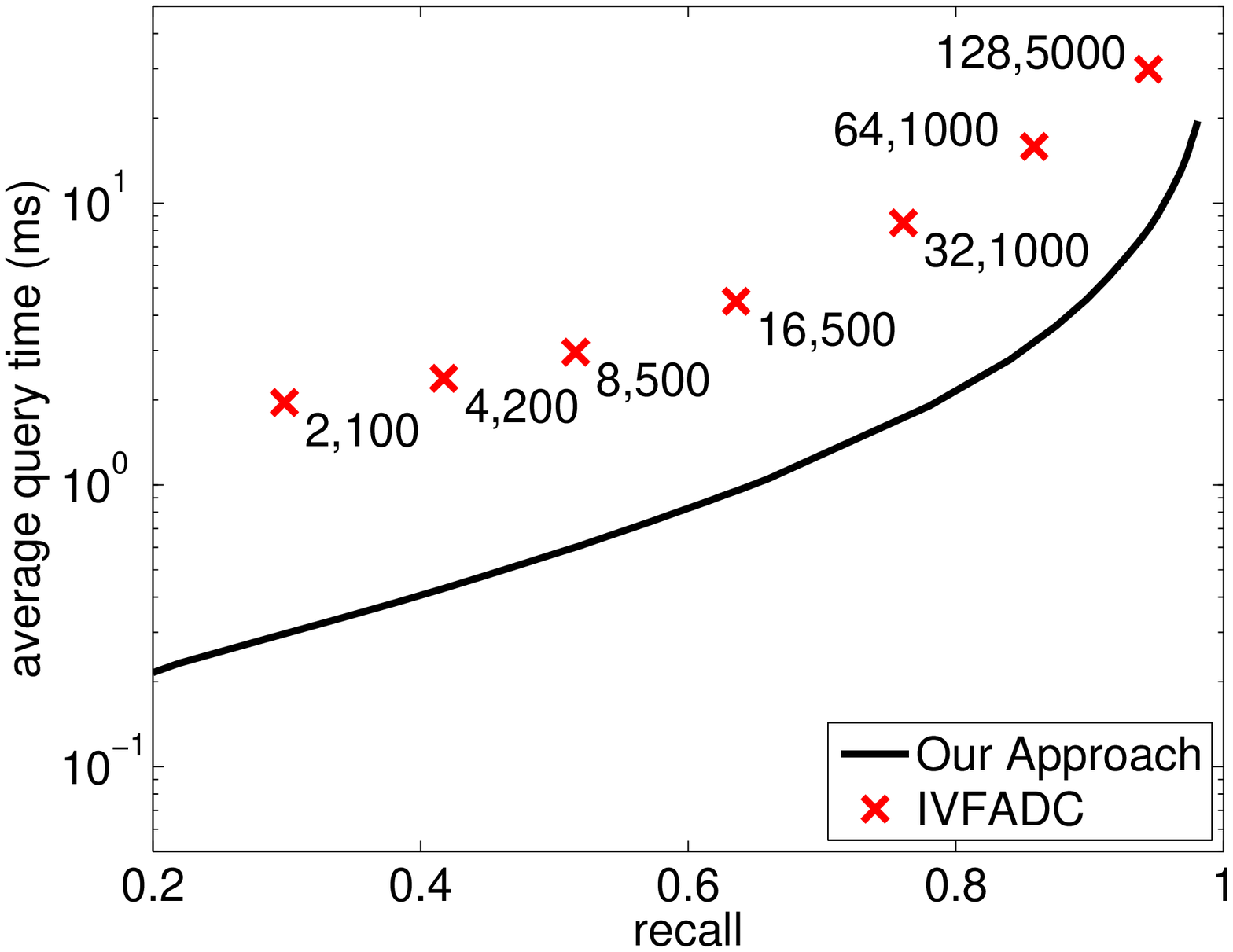}}
\caption{Search performances comparison with IVFADC~\cite{JegouDS11}
over (a) $1M$ SIFT and (b) $1M$ GIST. The parameters $M$ (the number of inverted lists visited),
$L$ (the number of candidates for re-ranking)
are given beside each marker of IVFADC}
\label{fig:ExpPami}
\end{figure}

In addition,
we also conduct experiments
to compare the source coding based ANN search algorithm~\cite{JegouDS11}.
This algorithm compresses each data vector into a short code
using product quantization,
resulting in
the fast approximate distance computation between vectors.
We report the results from the IVFADC system that performs the best
as pointed in~\cite{JegouDS11}
over the $1$M SIFT and GIST features.
To compare IVFADC with our approach,
we follow the scheme in~\cite{JegouDS11}
to add a verification stage to the IVFADC system.
We cluster the data points into $K$
inverted lists
and use a $64$-bits code to represent each vector
as done in~\cite{JegouDS11}.
Given a query, we first find its $M$ nearest inverted lists,
then compute the approximate distance from the query
to each of the candidates in the retrieved inverted lists.
Finally we re-rank the top $L$ candidates using Euclidean distance
and compute the $1$-recall~\cite{JegouDS11} of the nearest neighbor
(the same to the definition
of the search accuracy for $1$-NN).
Experimental results show
that $K=2048$ gets superior performance.
Figure~\ref{fig:ExpPami}
shows the results
with respect to the parameters $M$ and $L$.
One can see that our approach gets superior performance.

\subsection{Experiments over the BIGANN dataset}
We evaluate the performance of our approach
when combining it with the IVFADC system~\cite{JegouDS11}
for searching  very large scale datasets.
The IVFADC system organizes the data
using inverted indices built via a coarse quantizer
and represents each vector
by a short code produced by product quantization.
During the search stage,
the system visits the inverted lists
in ascending order of the distances to the query
and re-ranks the candidates according to the short codes.
The original implementation
only uses a small number of inverted lists
to avoid the expensive time cost in finding the exact nearest inverted indices.
The inverted multi-index~\cite{BabenkoL12}
is used to replace the inverted indices
in the IVFADC system,
which is shown better than the original IVFADC implementation~\cite{JegouDS11}.

We propose to
replace the nearest inverted list identification
using our approach.
The good search quality of our approach in terms of both accuracy and efficiency
makes it feasible to handle a large number of inverted lists.
We quantize the $1B$ features
into millions ($6M$ in our implementation) of groups
using a fast approximate k-means clustering algorithm~\cite{WangWKZL12},
and compute the centers of all the groups
forming the vocabulary.
Then we use our approach
to assign each vector to the inverted list
corresponding to the nearest center,
producing the inverted indices.
The residual displacement between each vector and its center
is quantized using product quantization
to obtain extra bytes for re-ranking.
During the search stage,
we find the nearest inverted lists to the query
using our approach
and then do the same reranking procedure
as in~\cite{BabenkoL12, JegouDS11}

Following~\cite{BabenkoL12, JegouDS11}
we calculate the recall@$T$ scores of the nearest neighbor
with respect to different length of the visited candidate list $L$
and different numbers of extra bytes, $m = 8, 16$.
The recall@$T$ score is equivalent to the accuracy
for the nearest neighbor
if a short list of $T$ vectors is verified
using exact Euclidean distances~\cite{JegouTDA11}.
The performance is summarized in Table~\ref{tab:bigann}.
It can be seen that our approach
consistently outperforms Multi-D-ADC~\cite{BabenkoL12}
and IVFADC~\cite{JegouDS11}
in terms of both recall and time cost
when retrieving the same number of visited candidates.
The superiority over IVFADC
stems from
that our approach significantly increases the number of inverted indices
and produces space partitions with smaller (coarse) quantization errors
and that our system accesses a few coarse centers
while guarantees relatively accurate inverted lists.
For inverted multi-index approach,
although the total number of centers is quite large
the data vectors are not evenly divided into inverted lists.
As reported in the supplementary material of~\cite{BabenkoL12},
$61\%$ of the inverted lists are empty.
Thus the quantization quality is not as good as ours.
Consequently, it performs worse than our approach.

\begin{table}[t]
\sidecaption[t]
\begin{tabular}[b]{|cc|ccc|c|}
\hline
System & List len. & R@$1$ & R@$10$ & R@$100$ & Time \\
\hline
\multicolumn{6}{|c|}{BIGANN, $1$ billion SIFTs, $8$ bytes per vector} \\
\hline
IVFADC & $4$ million & $0.100$ & $0.280$ & $0.600$ & $960$ \\
\hline
Multi-D-ADC & $10000$ & $0.165$ & $0.492$ & $0.726$ & $29$ \\
Multi-D-ADC & $30000$ & $0.172$ & $0.526$ & $0.824$ & $44$ \\
Multi-D-ADC & $100000$ & $0.173$ & $0.536$ & $0.870$ & $98$ \\
\hline
Graph-D-ADC & $10000$ & $0.199$ & $0.562$ & $0.802$ & $24$ \\
Graph-D-ADC & $30000$ & $0.201$ & $0.584$ & $0.873$ & $39$ \\
Graph-D-ADC & $100000$ & $0.201$ & $0.589$ & $0.896$ & $90$ \\
\hline
\hline
\multicolumn{6}{|c|}{BIGANN, $1$ billion SIFTs, $16$ bytes per vector} \\
\hline
IVFADC & $4$ million & $0.220$ & $0.610$ & $0.890$ & $1135$ \\
\hline
Multi-D-ADC & $10000$ & $0.324$ & $0.685$ & $0.755$ & $30$ \\
Multi-D-ADC & $30000$ & $0.347$ & $0.777$ & $0.891$ & $47$ \\
Multi-D-ADC & $100000$ & $0.354$ & $0.813$ & $0.959$ & $109$ \\
\hline
Graph-D-ADC & $10000$ & $0.374$ & $0.764$ & $0.831$ & $24$ \\
Graph-D-ADC & $30000$ & $0.391$ & $0.829$ & $0.924$ & $39$ \\
Graph-D-ADC & $100000$ & $0.395$ & $0.851$ & $0.964$ & $92$ \\
\hline
\end{tabular}
\caption{The performance
(recall for the top-$1$, top-$10$, and top-$100$ candidates after reranking
and average search time in milliseconds)
comparison between IVFADC~\cite{JegouTDA11}, Multi-D-ADC~\cite{BabenkoL12}
and Our approach (Graph-D-ADC).
IVFADC uses inverted lists with $K = 1024$,
Multi-D-ADC uses the second-order multi-index with $K = 2^{14}$
and our approach use inverted lists with $K = 6M$}
\label{tab:bigann}
\end{table}

\section{Analysis and discussion}
\label{sec:discussion}
\mytextbf{Index structure size.}
In addition to the neighborhood graph
and the reference vectors,
the index structure of our approach
includes a bridge graph and the bridge vectors.
The number of bridge vectors in our implementation
is $O(N)$,
with $N$ being the number of the reference vectors.
The storage cost of the bridge vectors
are then $O(\sqrt[m]{N})$,
and the cost of the bridge graph is also $O(N)$.
In the case of $1M$ $384$-dimensional GIST byte-valued features,
without optimization,
the storage complexity ($125M$ bytes) of the bridge graph
is smaller than the reference vectors ($384M$ bytes)
and the neighborhood graph ($160M$ bytes).
The cost of KD trees, VP trees,
and TP trees are $\sim$$180M$,
$\sim$$180M$, and $\sim$$560M$ bytes.
In summary,
the storage cost of our index structure
is comparable with those neighborhood graph
and tree-based structures.

In comparison to source coding~\cite{JegouDS11, JegouTDA11} and hashing
without using the original features,
and inverted indices (e.g.~\cite{BabenkoL12}),
our approach takes more storage cost.
However, the search quality of our approach
in terms of accuracy and time
is much better,
which leaves users for algorithm selection
according to their preferences
to less memory or less time.
Moreover the storage costs for $1M$ GIST and SIFT features ($<1G$ bytes)
and even $10M$ HOG features ($<8G$ bytes)
are acceptable in most today's machines.
When applying our approach to the BIGANN dataset of
$1B$ SIFT features,
the index structure size for our approach is about $14G$ for $m = 8$
and $22G$ for $m = 16$,
which is similar with Multi-D-ADC~\cite{BabenkoL12}
($13G$ for $m = 8$ and $21G$ for $m = 16$)
and IVFADC~\cite{JegouDS11}
($12G$ for $m = 8$ and $20G$ for $m = 16$).

\mytextbf{Construction complexity.}
The most time-consuming process
in constructing the index structure in our approach
is the construction of the neighborhood graph.
Recent research~\cite{WangWZTGL12}
shows that an approximate neighborhood graph can be built
in $O(N\log N)$ time,
which is comparable to the cost of constructing the bridge graph.
In our experiments, using a
$3.4G$ Hz quad core Intel PC,
the index structures of the $1M$ SIFT data,
the $1M$ GIST data, and the $10M$ HOG data
can be built within half an hour, an hour, and $10$ hours, respectively.
These time costs are relatively large
but acceptable as they are offline processes.

The algorithm of combining our approach with the IVFADC system~\cite{JegouDS11}
over the BIGANN dataset of size $1$ billion
requires the similar construction cost
with the state-of-the-art algorithm~\cite{BabenkoL12}.
Because the number of data vectors is very large ($1B$),
the most time-consuming stage is to assign each vector
to the inverted lists
and both take about $2$ days.
The structure of our approach organizing the $6M$ centers
takes only a few hours, which is relatively small.
These construction stages are all run
with $48$ threads
on a server with $12$ AMD Opteron $1.9G$Hz quad core processors.

\mytextbf{Search complexity.}
The search procedure of our approach consists of
the distance computation over the subvectors,
the traversal over the bridge graph
and the neighborhood graph.
The distance computation over the subvectors is very cheap
and takes small constant time
(about the
distance computation cost with $100$ vectors in our experiments).
Compared with the number of reference vectors
that are required to reach an acceptable accuracy
(e.g., the number is about $4800$ for accuracy $90\%$
in the $1M$ $384$-dimensional GIST feature data set),
such time cost is negligible.

Besides the computation of the distances
between the query vector and the visited reference vectors,
the additional time overhead comes
from maintaining the priority queue
and querying the bridge vectors
using the multi-sequence algorithm.
Given there are $T$ reference vectors that have been discovered,
it can be easily shown that the main queue is no longer than $T$.
Consider the worst case that all the $T$ reference vectors come from the bridge graph,
where each bridge vector is associated with $\alpha$ unique reference vectors on average
(the statistics for $\alpha$ in our experiments
is presented in Table~\ref{tab:parameter}),
we have that $\frac{T}{\alpha}$ bridge vectors are visited.
Thus, the maintenance of the main queue
takes $O((1+\frac{1}{\alpha})T\log T)$ time.
Extracting $\frac{T}{\alpha}$ bridge vectors using the multi-sequence algorithm~\cite{BabenkoL12}
takes $O(\frac{T}{\alpha}\log(\frac{T}{\alpha}))$.
Consequently
the time overhead on average is $O((1+\frac{2}{\alpha})T\log T - \frac{T}{\alpha}\log\alpha) = O(T\log T)$.

\begin{figure}[t]
\centering
\includegraphics[width = .9\linewidth, clip]{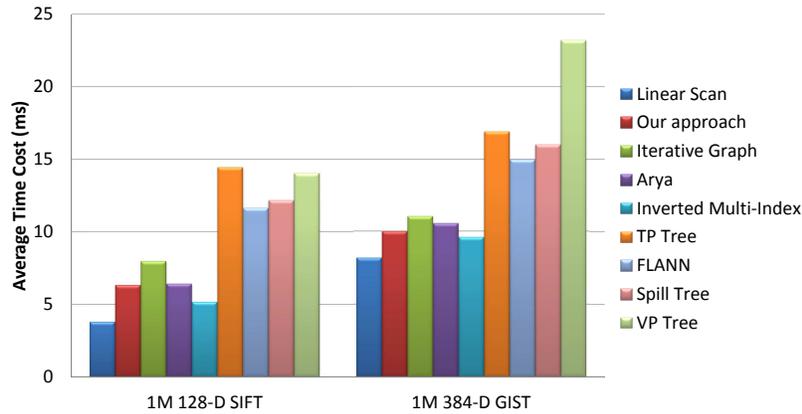}
\caption{Average time cost of visiting $10K$ reference vectors.
The time overhead (difference between the average time cost and the cost of liner scan) of our approach
is comparably small
 }\label{fig:overheadtime}
\end{figure}

Figure~\ref{fig:overheadtime} shows
the time cost of visiting $10K$ reference vectors
in different algorithms on two datasets.
Linear scan represents
the time cost of computing the distances
between a query and all reference vectors.
The overhead of a method
is the difference between the time cost
of this method
and that of linear scan.
We can see that the inverted multi-index takes the minimum overhead
and our approach is the second minimum.
This is because our approach includes
extra operations over the main queue.

\mytextbf{Relations to~source coding~\cite{JegouDS11}
and inverted multi-index~\cite{BabenkoL12}.}
Product quantization (or generally Cartesian concatenation)
has two attractive properties.
One property is that it is able to produce a large set of concatenated vectors
from several small sets of subvectors.
The other property is that
the exact nearest vectors to a query vector from such a large set of concatenated vectors
can be quickly found using the multi-sequence algorithm.
The application to source coding~\cite{JegouDS11}
exploits the first property,
thus results in fast distance approximation.
The application to inverted multi-index~\cite{BabenkoL12}
makes use of the second property
to fast retrieve concatenated quantizers.
In contrast,
our approach exploits both the two properties:
the first property guarantees that
the approximation error of the concatenated vectors to the reference vectors
is small with small sets of subvectors,
and the second property guarantees that
the retrieval from the concatenated vectors is very efficient
and hence the time overhead is small.

\section{Conclusions}
The key factors contribute to the superior performance of our proposed approach
include:
(1) Discovering NN candidates from the neighborhood
of both bridge vectors and reference vectors
is very cheap;
(2) The NN candidates from the neighborhood of the bridge vector
have high probability to be true NNs
because there are a large number of effective bridge vectors
generated by Cartesian concatenation;
(3) Retrieving nearest bridge vectors is very efficient.
The algorithm is very simple
and is easily implemented.
The power of our algorithm is demonstrated
by the superior ANN search performance
over large scale SIFT, HOG, and GIST datasets,
as well as
over a very large scale dataset,
the BIGANN dataset of $1$ billion SIFT features
through the combination of
our approach
with the IVFADC system.

{\small
\bibliographystyle{./styles/spmpsci}
\bibliography{bow}
}
\end{document}